# Error-tolerant Tree Matching

Kemal Oflazer
Department of Computer Engineering and Information Science,
Bilkent University, Ankara, TR-06533, Turkey
ko@cs.bilkent.edu.tr

## Abstract

This paper presents an efficient algorithm for retrieving from a database of trees, all trees that match a given query tree *approximately*, that is, within a certain error tolerance. It has natural language processing applications in searching for matches in example-based translation systems, and retrieval from lexical databases containing entries of complex feature structures. The algorithm has been implemented on SparcStations, and for large randomly generated synthetic tree databases (some having tens of thousands of trees) it can associatively search for trees with a small error, in a matter of tenths of a second to few seconds.

## 1 Introduction

Recent approaches in machine translation known as example-based translation rely on searching a database of previous translations of sentences or fragments of sentences, and composing a translation from the translations of any matching examples (Sato and Nagao, 1990; Nirenburg, Beale and Domasnhev, 1994). The example database may consist of paired text fragments, or trees as in Sato and Nagao (1990). Most often, exact matches for new sentences or fragments will not be in the database, and one has to consider examples that are "similar" to the sentence or fragment in question. This involves associatively searching through the database, for trees that are "close" to the query tree. This paper addresses the computational problem of retrieving trees that are close to a given query tree in terms of a certain distance metric.

The paper first presents the approximate tree matching problem in an abstract setting and presents an algorithm for approximate associative tree matching. The algorithm relies on linearizing the trees and then representing the complete database of trees as a trie structure which can be efficiently searched. The problem then reduces to sequence correction problem akin to standard spelling correction problem. The trie is then used with an approximate finite state recognition algorithm close to a query tree. Following some experimental results from a number of synthetic tree databases, the paper ends with conclusions.

## 2 Approximate Tree Matching

In this paper we consider the problem of searching in a database of trees, all trees that are "close" to a given query tree, where closeness is defined in terms of an error metric. The trees that we consider have labeled terminal and non-terminal nodes. We assume that all immediate children of a given node have unique labels, and that a total ordering on these labels is defined. We consider two trees close if we can

- add/delete a small number of leaves to/from one of the trees, and/or
- change the label of a small number of leaves in one of the trees

to get the second tree. A pair of such "close" trees is depicted in Figure 1.

### 2.1 Linearization of trees

Before proceeding any further we would like to define the terminology we will be using in the following sections: We identify each leaf node in a tree with an ordered *vertex list* $(v_0, v_1, v_2, \cdots, v_d)$ where each $v_i$ is the label of a vertex from the root $v_0$ to the leaf $v_d$ at depth $d$, and for $i > 0$, $v_i$ is the parent of $v_{i+1}$. A tree with $n$ leaves is represented by a *vertex list sequence* $VLS = V_1, V_2, \cdots, V_n$ where each $V_j = v_0^j, v_1^j, v_2^j, v_3^j, \cdots, v_{d_j}^j$ corresponds to a vertex list for a leaf at level $d_j$. This sequence is constructed by taking into account the total order on the labels at every level, that is, $V_i$ is *lexicographically less than* $V_{i+1}$, based on the total ordering of the vertex labels. For instance, the first tree in Figure 1 would be represented by the vertex list sequence:

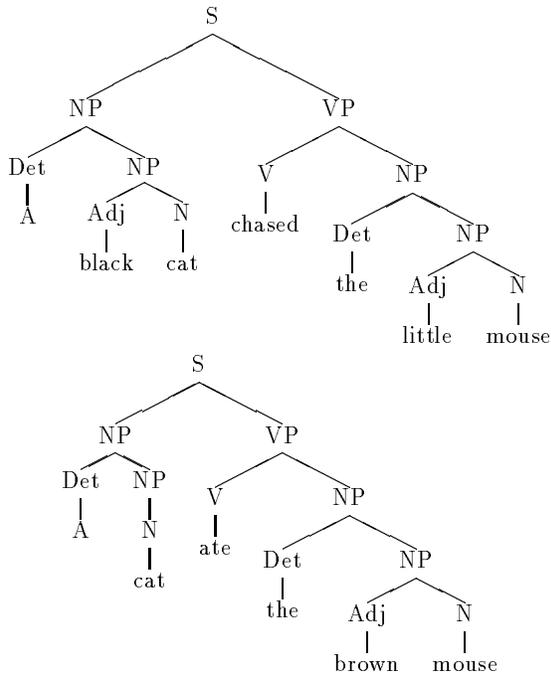
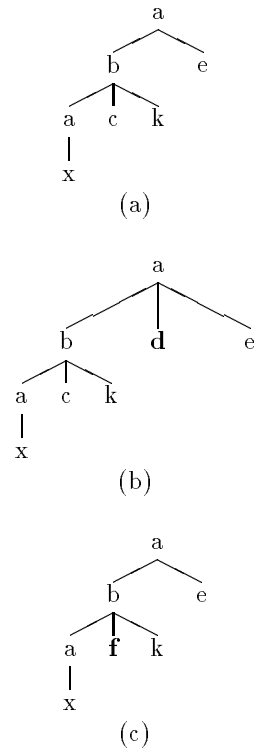

Figure 1: Trees that are "close" to each other.

```
((S,NP,Det,a),
 (S,NP,NP,Adj,black),
 (S,NP,NP,N,cat),
 (S,VP,NP,Det,the),
 (S,VP,NP,NP,Adj,little),
 (S,VP,NP,NP,N, mouse),
 (S,VP,V,chased))
```

assuming the normal lexicographic ordering on node names.

### 2.2 Distance between two trees

We define the distance between two trees according to the *structural differences* or *differences in leaf labels*. We consider an extra or a missing leaf as a structural change. If, however, both trees have a leaves whose vertex lists match in all but the last (leaf vertex) label, we consider this as a difference in leaf labels. For instance, in Figure 2, there is extra leaf in tree (b) in comparison to the tree in (a), while tree (c) has a leaf label difference. We associate the following costs associated with these differences:

- If both trees have a leaf whose vertex list matches in all but the last (leaf vertex) label, we assign a label difference error of $C$.

- If a certain leaf is missing in one of the trees but exists in the other one, then we assign a cost $S$ for this a structural difference.

We currently treat all structural or leaf label differences as incurring a cost that is independent of the tree level at which the difference takes place.

Figure 2: Structural and leaf label differences between trees.

If, however, differences that are closer to the root of the tree are considered to be more serious than differences further away from the root, it is possible to modify the formulation to take this into account.

### 2.3 Converting a set of trees into a trie

A *tree database $D$* consists of a set of trees $T_1, T_2, \cdots, T_k$, each $T_i$ being a vertex list sequence for a tree. Once we convert all the trees to a linear form, we have a set of vertex list sequences. We can convert this set into a trie data structure. This trie will compress any possible redundancies in the prefixes of the vertex list sequences to achieve a certain compaction which helps during searching.[1]

For instance, the three trees in Figure 2 can be represented as a trie as shown in Figure 3. The edge labels along the path to a leaf when concatenated in order gives the vertex list sequence for a tree, e.g., `((a,b,a,x), (a,b,c), (a,b,k), (a,e))` represents the tree (a) in Figure 2.

---

[1] Note that it is possible to obtain more space reduction by also sharing any common postfixes of the vertex label sequences using a directed acyclic graph representation and not a trie, but this does not improve the execution time.

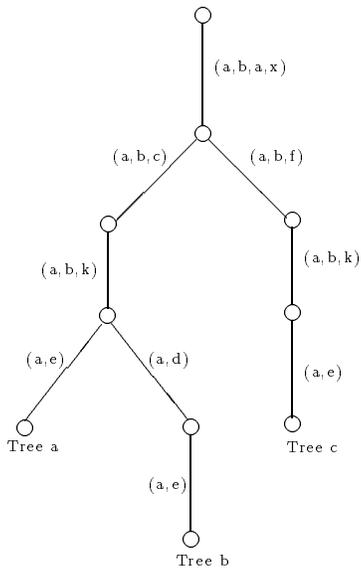

Figure 3: Trie representation of the 3 trees in Figure 2

### 2.4 Error-tolerant matching in the trie

Our concern in this work is not the exact match of trees but rather approximate match. Given the vertex list sequence for a query tree, exact match over the trie can be performed using the standard techniques by following the edge labeled with next vertex list until a leaf in the trie is reached, and the query vertex label sequence is exhausted. For approximate tree matching, we use the error-tolerant approximate finite-state recognition algorithm (Oflazer, 1996), which finds all strings within a given error threshold of some string in the regular set accepted by the underlying finite-state acceptor. An adaptation of this algorithm will be briefly summarized here.

Error-tolerant matching of vertex list sequences requires an error metric for measuring how much two such sequences deviate from each other. The *distance* between two sequences measures the minimum number of insertions, deletions and leaf label changes that are necessary to convert one tree into another. It should be noted that this is different from the error metric defined by (Wang et al., 1994).

Let $Z = Z_1, Z_2, \ldots, Z_p$, denote a generic vertex list sequence of $p$ vertex lists. $Z[j]$ denotes the initial subsequence of $Z$ up to and including the $j^{th}$ leaf label. We will use $X$ (of length $m$) to denote the query vertex list sequence, and $Y$ (of length $n$) to denote the sequence that is a (possibly partial) candidate vertex list sequence (from the database of trees).

Given two vertex list sequences $X$ and $Y$, the distance, $dist(X[m], Y[n])$, computed according to the recurrence below, gives the minimum number of leaf insertions, deletions or leaf label changes necessary to change one tree to the other.

$$dist(X[m], Y[n]) = dist(X[m-1], Y[n-1])$$

if $x_m = y_n$
(last vertex lists are same)

$$= dist(X[m-1], Y[n-1]) + C$$

if $x_m$ and $y_n$
differ only at the
leaf label

$$= dist(X[m-1], Y[n]) + S$$

if $y_n < x_m$ (lexicographically)
$X$ is missing leaf $y_n$.

$$= dist(X[m], Y[n-1]) + S$$

if $x_m < y_n$ (lexicographically)
$X$ has an extra leaf $x_m$.

Boundary Conditions
$$dist(X[0], Y[n]) = n \cdot S$$
$$dist(X[m], Y[0]) = m \cdot S$$

For a tree database $D$ and a distance threshold $t > 0$, we consider a query tree represented by a vertex list sequence $X[m]$ (not in the database) to match the database with an error of $t$, if the set

$$C = \{Y[n] \mid Y[n] \in D \text{ and } dist X[m], Y[n]) \leq t\}$$

is not empty.

### 2.5 An algorithm for approximate tree matching

Standard searching with a trie corresponds to traversing a path starting from the start node (of the trie), to one of the leaf nodes (of the trie), so that the concatenation of the labels on the arcs along this path matches the input vertex list sequence. For error-tolerant matching, one needs to find <u>*all paths from the start node to one of the final nodes, so that when the labels on the edges along a path are concatenated, the resulting vertex list sequence is within a given distance threshold $t$, of the query vertex list sequence.*</u>

This search has to be very fast if approximate matching is to be of any practical use. This means that paths in the trie that can lead to no solutions have to be pruned so that the search can be limited to a very small percentage of the search space. We need to make sure that any candidate (prefix) vertex list sequence that is generated as the search is being performed, does not deviate from certain initial subsequences of the query sequence by more than the allowed threshold. To detect such cases, we use the notion of a *cut-off distance*. The cut-off distance measures the minimum distance between an initial subsequence of the query sequence sequence, and the (possibly partial) candidate sequence. Let $Y$ be a partial candidate sequence whose length is $n$, and let $X$ be the query sequence of length $m$. Let $l = \min(1, n - \lfloor t/M \rfloor)$ and $u = \max(m, n + \lceil t/M \rceil)$ where $M$ is the cost of insertions and deletions. The cut-off distance $cutdist(X[m], Y[n])$ is defined as

$$cutdist(X[m], Y[n]) = \min_{l \leq i \leq u} dist(X[i], Y[n]).$$

Note that except at the boundaries, the initial subsequences of the query sequence $X$ considered are of length $n - \lfloor t/M \rfloor$ to length $n + \lceil t/M \rceil$. Any initial subsequence of $X$ shorter than $l$ needs more than $\lfloor t/M \rfloor$ leaf node insertions, and any initial substring of $X$ longer than $u$ requires more than $+\lceil t/M \rceil$ leaf node deletions, to at least equal $Y$ in length, violating the distance constraint.

Given a vertex list sequence $X$ (corresponding to a query tree), a partial candidate sequence $Y$ is generated by successively concatenating labels along the arcs as transitions are made, starting with the start state. Whenever we extend $Y$ going along the trie, we check if the cut-off distance of $X$ and the partial $Y$ is within the bound specified by the threshold $t$. If the cut-off distance goes beyond the threshold, the last edge is backed off to the source node (in parallel with the shortening of $Y$) and some other edge is tried. Backtracking is recursively applied when the search can not be continued from that node. If, during the construction of $Y$, a terminal node (which may or may not be a leaf of the trie) is reached without violating the cutoff distance constraint, and $dist(X[m], Y[n]) \leq t$ at that point, then $Y$ is a tree in the database that matches the input query sequence.[2]

Denoting the nodes of the trie by subscripted $q$'s ($q_0$ being the initial node (e.g., top node in Figure 3)) and the labels of the edges by $V$, and denoting by $\delta(q_i, V)$ the node in the trie that one can reach from node $q_i$ with edge label $V$ (denoting a vertex list), we present, in Figure 4, the algorithm for generating all $Y$'s by a (slightly modified) depth-first probing of the trie. The crucial point in this algorithm is that the cut-off distance computation can be performed very efficiently by maintaining a matrix $H$ which is an $m$ by $n$ matrix with element $H(i, j) = dist(X[i], Y[j])$ (Du and Chang, 1992). We can note that the computation of the element $H(i+1, j+1)$ recursively depends on only $H(i, j), H(i, j+1), H(i+1, j)$ from the earlier definition of the edit distance (see Figure 5.) During the depth first search of the trie, entries in column $n$ of the matrix $H$ have to be (re)computed, only when the candidate string is of length $n$. During backtracking, the entries for the last column are

[2] Note that we have to do this check since we may come to other irrelevant terminal nodes during the search.

```
/*push empty candidate, and start
  node to start search */
push((ε, q_0))
while stack not empty
begin
  pop((Y', q_i)) /* pop partial sequence Y'
                   and the node */
  for all q_j and V such that δ(q_i, V) = q_j
    begin /* extend the candidate sequence */
      Y = concat(Y', V)
          /* n is the current length of Y */
    /* check if Y has deviated too much,
       if not push */
      if cutdist(X[m], Y[n]) ≤ t then push((Y, q_j))
    /* also see if we are at a final state */
      if dist(X[m], Y[n]) ≤ t and
         q_j is a terminal node then output Y
    end
end
```

Figure 4: Algorithm for error-tolerant recognition of vertex list sequences

$$\begin{pmatrix} \cdots & \cdots & \cdots & \cdots & \cdots \\ \vdots & \vdots & \vdots & \vdots & \vdots \\ \cdots & \cdots & \cdots & \cdots & \cdots \\ \cdots & \cdots & H(i,j) & H(i,j+1) & \cdots \\ \cdots & \cdots & H(i+1,j) & H(i+1,j+1) & \cdots \\ \vdots & \vdots & \vdots & \vdots & \vdots \\ \cdots & \cdots & \cdots & \cdots & \cdots \end{pmatrix}$$

Figure 5: Computation of the elements of the $H$ matrix.

discarded, but the entries in prior columns are still valid. Thus all entries required by $H(i+1, j+1)$, except $H(i, j+1)$, are already available in the matrix in columns $i - 1$ and $i$. The computation of $cutdist(X[m], Y[n])$ involves a loop in which the minimum is computed. This loop (indexing along column $j + 1$) computes $H(i, j + 1)$ before it is needed for the computation of $H(i + 1, j + 1)$.

## 3 Experimental Results

We have experimented with 3 syntheticly generated sets of trees with the properties given in Table 1. In this table, the third column (label ALP) gives the average ratio of the vertices at each level which are randomly selected as leaf vertices in a tree. The fourth column gives the maximum number of children that a non-leaf node may have. The last column gives the maximum depth of the trees in that database.

From these synthetic databases, we randomly extracted 100 trees and then perturbed them with random leaf deletions, insertions and label changes so that they were of some distance from a tree in the original tree. We used thresholds $t = 2$ and $t = 4$, allowing an error of $C = 1$ for each leaf label change and an error of $S = 2$ for each insertion or deletion (see Section 2.2). We then ran our algorithm on these data sets and obtained performance information. All runs were performed on a Sun SparcStation 20/61 with 128M real memory. The results are presented in Table 2. It can be seen that the approximate search algorithm is very fast for the set of synthetic tree databases that we have experimented with. It certainly is also possible that additional space savings can be achieved if directed acyclic graphs can be used to represent the tree database taking into account both common prefixes and common suffixes of vertex list sequences.

| Database | Number of Trees | ALP | Max Children | Max Depth |
|---|---|---|---|---|
| 1 | 1,000 | 1/3 | 8 | 5 |
| 2 | 10,000 | 1/2 | 16 | 5 |
| 3 | 50,000 | 1/2 | 8 | 3 |

Table 1: Properties of the synthetic databases of trees

| Database | Threshold | Avg. Leaves/ Query Tree | Avg. Search Time (Msec) | Avg. Trees Found/ Query |
|---|---|---|---|---|
| 1 | 2 | 12.00 | 65 | 1.96 |
|   | 4 | 12.42 | 81 | 16.65 |
| 2 | 2 | 24.65 | 990 | 3.32 |
|   | 4 | 25.62 | 1,659 | 31.59 |
| 3 | 2 | 10.45 | 2,550 | 13.63 |
|   | 4 | 10.45 | 3,492 | 68.62 |

Table 2: Performance results for the approximate tree matching algorithm.

## 4 Conclusions

This paper has presented an algorithm for approximate associative tree matching that can be used in example-based machine translation applications. The algorithm efficiently searches in a database of trees, all trees that are "close" to a given query tree. The algorithm has been implemented on Sun Sparcstations, and experiments on rather large synthetic tree database indicate that it can perform approximate matches within tenths of a second to few seconds depending on the size of the database and the error that the search is allowed to consider.

## 5 Acknowledgments

This research was in part funded by a NATO Science for Stability Phase III Project Grant – TU-LANGUAGE.